\documentclass[lettersize,journal]{IEEEtran}
\usepackage{amsmath,amsfonts}
\usepackage{algorithmic}
\usepackage{algorithm}
\usepackage{array}
\usepackage{textcomp}
\usepackage{stfloats}
\usepackage{url}
\usepackage{verbatim}
\usepackage{graphicx}
\usepackage{cite}
\usepackage{amsmath}
\usepackage{amssymb}
\usepackage{mathtools}
\usepackage{amsthm}
\usepackage{booktabs}
\usepackage{subcaption}
\usepackage{placeins}

\newcommand{\ieno}{\textit{i}.\textit{e}.}
\newcommand{\egno}{\textit{e}.\textit{g}.} 
\newcommand{\etal}{\textit{et al.}}
\newcommand{\mywidth}{0.9}
\newcommand{\ours}{\emph{R-VLM}}

\usepackage{xcolor}
\usepackage{xr}
\usepackage{tabularx}

\newcommand{\bigstrut}{\rule[-0.3\baselineskip]{0pt}{0.7\baselineskip}}

\begin{document}

\title{Long Video Understanding with Learnable Retrieval in Video-Language Models}

\author{Jiaqi Xu,
        Cuiling Lan, 
        Wenxuan Xie,
        Xuejin Chen,
        Yan Lu
\thanks{J. Xu is with the School of Information Science and Technology, University of Science and Technology of China, Hefei 230026, China. E-mail: xujiaqi@mail.ustc.edu.cn}
\thanks{C. Lan, W. Xie and Y. Lu are with the Microsoft Research Asia, Beijing 100080‌, China. E-mail: \{culan, wenxie, yanlu\}@microsoft.com}
\thanks{X. Chen is with the School of Information Science and Technology, University of Science and Technology of China, Hefei 230026, China. E-mail: xjchen99@ustc.edu.cn}
\thanks{This work was done when J. Xu was an intern at Microsoft Research Asia. This paper is the result of an open source research project.}
\thanks{This paper has supplementary downloadable material available at http://ieeexplore.ieee.org., provided by the author. The material includes more details and experimental results.}}



\maketitle

\begin{abstract}
The remarkable natural language understanding, reasoning, and generation capabilities of large language models (LLMs) have made them attractive for application to video understanding, utilizing video tokens as contextual input. However, employing LLMs for long video understanding presents significant challenges. The extensive number of video tokens leads to considerable computational costs for LLMs while using aggregated tokens results in loss of vision details. Moreover, the presence of abundant question-irrelevant tokens introduces noise to the video reasoning process. To address these issues, we introduce a simple yet effective learnable retrieval-based video-language model (R-VLM) for efficient long video understanding. Specifically, given a question and a long video, our model identifies the most relevant $K$ video chunks and uses their associated visual tokens to serve as context for the LLM inference. This effectively reduces the number of video tokens, eliminates noise interference, and enhances system performance. We achieve this by incorporating a learnable lightweight MLP block to facilitate the efficient retrieval of question-relevant chunks, through the end-to-end training of our video-language model with a proposed soft matching loss. Experimental results on multiple zero-shot video question answering datasets validate the effectiveness of our framework for comprehending long videos. 
\end{abstract}


\section{Introduction}
\label{sec:Introduction}

\IEEEPARstart{W}{ith} the rapid development of the Internet and the widespread use of cameras and smartphones, both individuals and businesses are generating massive amounts of video data every day in various fields such as entertainment, education, and technology. In such era of information explosion, understanding and extracting information from video content has become increasingly important to better meet people's needs and promote social progress. In this context, Video Question Answering (video QA)
emerges as an essential interface for extracting and delivering information from video content. Video QA systems allow users to ask natural language questions about videos and receive answers based on the visual (and auditory) information within the video.

There is a growing trend towards leveraging LLMs for video QA \cite{maaz2023video, wang2023chatvideo, li2023videochat,zhang2023video, luo2024video}. On one hand, LLMs benefit from the vast knowledge acquired through training on enormous text corpora; on the other hand, they provide users with a more natural and intuitive way to interact with video data. 
Generally, visual tokens extracted from a video snippet are transformed and used as input (prompt) to the LLM, along with the text query. 
However, it is worth noting that the consumption of abundant visual tokens by a LLM can significantly increase the memory and computational burden, making it unaffordable for low-resource GPU agents. To mitigate this issue,
Video-ChatGPT \cite{maaz2023video} performs global spatial and temporal pooling on the video tokens, although this comes at the cost of losing detail due to pooling.
Video-LLaMA \cite{zhang2023video} aggregates video tokens using a Q-former with cross attention. Most of these methods are designed for short-video QA tasks, where answer-related frames usually spread over the trimmed video snippet.

\begin{figure*}[!t]
	\centering	\includegraphics[width=1.0\textwidth]{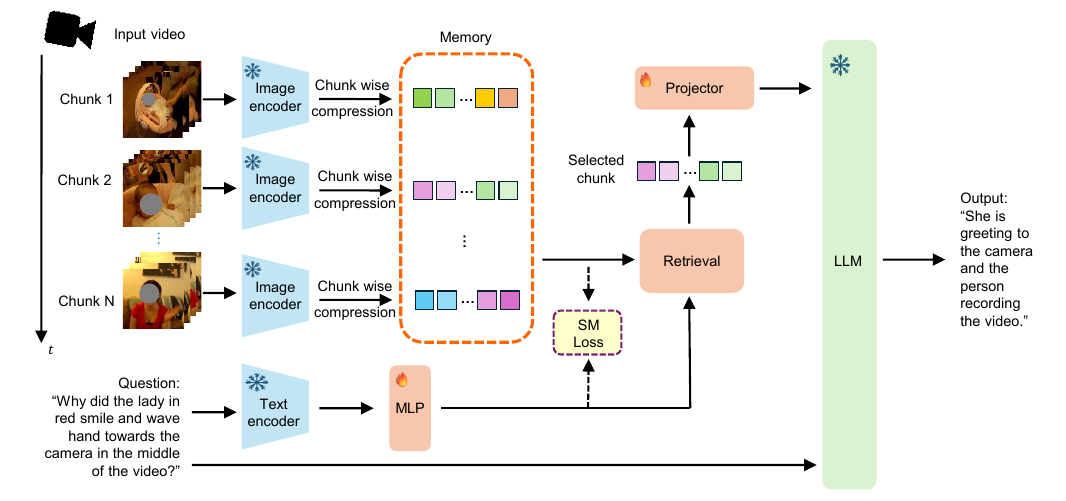}
    \caption{Illustration of our learnable retrieval-based video-language model for efficient long video question answering. We encode an input long video into a sequence of video chunks, with each chunk represented by a set of spatial and temporal visual tokens. Question-guided retrieval is performed to find the top $K$ relevant video chunks, with their tokens as the input to the LLM for answer generation. Here, we use $K=1$ for illustration purpose. A learnable lightweight MLP block (following the text encoder) and the projector are trained end-to-end, where the encoders and LLM are frozen. Soft matching (SM) loss is introduced to regularize the retrieval related learning.}
    \label{fig: framework}
\end{figure*}

In practical application scenarios, users often raise questions about long videos (\egno, longer than 1 minute), in which the segments
that are relevant to the questions usually constitute only a small fraction of the entire video.
The presence of answer-irrelevant segments constitutes redundancy and can potentially interfere with the video QA process, thus reducing its effectiveness.
Therefore, it is imperative to develop a simple yet efficient framework for long video QA. 

To address these challenges, we draw inspiration from biology and cognitive science. As we know, human working memory is a cognitive system responsible for temporarily holding and manipulating information necessary for complex tasks such as reasoning, learning, and comprehension \cite{thompson2013memory}. Faced with the vast amount of information stored in long-term memory, working memory selectively retrieves relevant information while filtering out irrelevant data for further cognition. Motivated by this, we aim to design a framework that is capable of identifying and concentrating on relevant video segments while filtering out irrelevant information, ensuring accurate and efficient question answering without imposing excessive computational demands.

In this paper, we propose a retrieval-based video-language model, R-VLM, for effective long-video question answering. Fig.~\ref{fig: framework} shows the overall framework. Specifically, given a long video and a question (query), we divide the video into a sequence of non-overlapping video chunks, with each chunk representing a short video segment (\egno, 4 seconds). Note that we sample at a low frame rate in considering the memory limitation and video redundancy. To allow a chunk to contain dynamic temporal information, we use 4 seconds that are expected to contain temporal dynamics as the chunk unit. We then aggregate the encoded tokens within a chunk through spatial and temporal pooling to obtain chunk tokens,  reducing redundancy while preserving considerable local details. We perform question-guided retrieval by introducing a lightweight \emph{learnable MLP block} on top of a CLIP text encoder to obtain the top $K$ most relevant chunks, and subsequently use the small set of tokens (after projection) from these chunks as input to the LLM for video reasoning. 
In this way, we efficiently match and pass the most question-pertinent visual information to the LLM for inference. 

Our framework demonstrates superior zero-shot generalization performance on multiple video QA datasets, 


Our contributions can be summarized below:
\begin{itemize}
\item We propose a learnable retrieval-based video-language model for effective long video understanding. 
\item We validate the feasibility of using learnable lightweight retrieval for selecting the most question-relevant video chunks to answer questions from a long video, within large video-language model in an end-to-end manner. 
\item Our method achieves superior performance and the ablation studies demonstrate the superiority of our retrieval mechanism. 
\end{itemize}

\section{Related Work}


Video Question Answering (Video QA) has emerged as a crucial task for extracting information from videos through natural language interactions. Before the prevalence of LLM, conventional methods focus on network and multi-modal interaction designs for video understanding \cite{yuan2020adversarial, 8811730, wang2021dualvgr, qian2023locate, zhang2023psam, guo2021universal, cheng2023keyword, jiang2022livlr}. Yuan \etal \cite{yuan2020adversarial} propose AMN to learn multi-modal feature representations by finding a more coherent subspace for video clips and the corresponding texts based on GANs. Wang \etal \cite{8811730} propose an alternating attention network, which iteratively focuses on frame regions, video frames, and words in the question in multi-turns to generate better joint representations of the video and question. 
DualVGR \cite{wang2021dualvgr} uses a multi-view graph attention network to provide a context-aware feature representation. A query punishment module is applied to modulate irrelevant visual features through multiple reasoning cycles to reduce their interference. Qian \etal \cite{qian2023locate} propose a scheme to localize the question to a segment in the video to infer the answer, where a 3DCNN network and attention modules are designed to extract features and generate location proposals. Based on existing LLM and visual encoder, how to design a simple framework that can leverage the LLM and visual encoder for long video understanding is still under-explored. We propose R-VLM that leverage a lightweight MLP block to enable a learnable retrieval module for efficient video question answering.

Recent advances in LLMs have inspired and promoted the integration of pre-trained LLMs with visual processing components for multimodal understanding \cite{alayrac2022flamingo}, \cite{li2023blip}, \cite{zhu2023minigpt}, \cite{liu2023visual}, \cite{huang2023language}, \cite{suris2023vipergpt}. Flamingo \cite{alayrac2022flamingo} and BLIP-2 \cite{li2023blip} are seminal works that leverage the pre-trained LLM and CLIP image encoder for zero-shot image-text abilities. BLIP2 \cite{li2023blip} introduces a Q-Former to map the image tokens from the CLIP encoder \cite{radford2021learning} to the textual embedding space of LLMs. InstructBLIP \cite{dai2305instructblip} uses instruction-aware Q-Former to globally aggregate the visual tokens. LLaVA \cite{liu2023visual} maps the image spatial tokens to the textual embedding space of LLMs using a linear projector. LLaVA \cite{liu2023visual}, MiniGPT4 \cite{zhu2023minigpt}, and mPLUG-owl \cite{ye2023mplug} promote instruction following using image-instruction-following datasets. 

Some works have focused on enabling video-language understanding by integrating vision encoder and LLMs \cite{damonlpsg2023videollama,maaz2023video,li2023videochat,wang2023chatvideo},\cite{weng2024longvlm}. Video-LLaMA \cite{damonlpsg2023videollama} uses a video Q-former to assemble the features from the pre-trained CLIP image encoder. Video-ChatGPT \cite{maaz2023video} performs spatial and temporal pooling of the feature maps encoded by the pre-trained CLIP image encoder. The ensembled visual tokens are taken as input to the LLMs. Most of these works globally aggregate video tokens. They may work well for short videos. But for the long video-language comprehension, they would be less efficient. The information of the question-related video segments may be submerged to the global-wise token representations, making the reasoning difficult. 

SeViLA \cite{yu2024self} uses the BLIP-2 image-language model to select temporal key frames which are subsequently inputted into BLIP-2 for video question answering. The process involves complex training and the frame selection is computationally expensive since the candidate frames must be processed by BLIP-2 (minimum 3.1B parameters), making it inefficient for long video understanding. 



In order to understand long videos, MovieChat \cite{song2024moviechat} leverages short- and long-term memories for video representation. LLaMA-VID \cite{li2023llama} reduces the computation and memory cost by representing each frame with two tokens, but still suffers from temporal redundancy and interference. EALD-MLLM \cite{li2024eald} addresses long video emotion analysis by segmenting videos into sampled clips, generating a description for each clip with Video-LLaMA, and leveraging ChatGPT to aggregate segment-level outputs into coherent global predictions. VILA \cite{lin2024vila} employs a Frame-Prompter for key-frame selection, but without question awareness, the selection is ineffective. SeViT \cite{kim2023semi} retrieves query-relevant frames for a text generator, yet its non-parametric frame selector prevents end-to-end optimization, leading to sub-optimal performance. Wang \etal \cite{wang2024weakly} fuse QA pairs as event descriptions to find key-frames as target moments and pseudo-labels. Unlike \cite{wang2024weakly}, we do not use pseudo-labels for supervision. VideoStreaming \cite{qian2024streaming} splits video into short clips, where a small language model (Phi-2, 2.7B) was used for clip selection.
VideoTree \cite{wang2024videotree} extracts query-relevant information from
the input video through an iterative process, based on their relevance to the query scored by an LLM. However, requiring an extra LLM adds complexity. 
In contrast, our method introduces a lightweight end-to-end trainable retrieval module with Soft Matching loss design for efficient video understanding. Such lightweight, efficient, and easy to integrate design is suitable for practical system deployment.

\section{Retrieval-based Video-Language Model}
\label{sec:method}

Given a lengthy video, it is time-consuming to watch the entire content to obtain the desired information. A video understanding system, which can automatically infer answers based on the long video and user's query (question), is in high demand. LLMs possess extensive world knowledge and reasoning capabilities, albeit at the expense of high computational complexity. Some previous works on multi-modal language models utilize a set of projected visual tokens as input (context) to LLM for inference \cite{li2023blip}, \cite{zhu2023minigpt}, where the inference cost is proportional to the number of input visual tokens.  

Leveraging powerful LLMs to understand long videos presents challenges. Firstly, we expect to use only a small set of tokens as input to reduce computational costs. Secondly, as the video length increases, in general, there is a corresponding growth in the number of visual tokens and the overall amount of information. Representing a long video with very few tokens becomes difficult. Furthermore, question-irrelevant information may interfere with reasoning. Motivated by the retrieval mechanism of brain, we introduce a question-guided retrieval mechanism to identify and select a few relevant video chunks as context of LLM.

Fig.~\ref{fig: framework} illustrates the overall framework of our proposed retrieval-based video-language model (R-VLM), designed for efficient long video understanding. The framework is comprised by several components: a frozen LLM, a frozen image encoder, a frozen text encoder, a memory for storing video chunk tokens, a retrieval module for selecting the question-relevant chunks, a learnable MLP block, and a projector. Through end-to-end training with cross-entropy loss and the proposed soft matching (SM) loss, the MLP block is optimized to learn to identify the most relevant $K$ chunks, while the projector is optimized to align the selected visual tokens with the text (question) space. 
In the following subsections, we provide a detailed explanation of the key components and designs.

\subsection{Video Tokenization}

Video tokenization involves encoding the raw video into visual tokens, which will be processed and then passed to the LLM for inference. We partition a long video into chunks, with each chunk represented by a set of compressed visual tokens. Such chunk tokens are stored in a memory to facilitate question-guided retrieval for efficient video question-answering.

\textbf{Chunking the video and feature extraction} Given a long video sample $V_i \in \mathbb{R}^{T_i\times H \times W \times 3}$ with $T_i$ frames, height $H$, and width $W$, we divide it into small non-overlapped video chunks (see Fig.~\ref{fig: framework}), which are the basic units with each containing spatial information and temporal dynamic for question-guided retrieval. We set the duration of a chunk to 4 seconds, \ieno, $M=4$ frames when frame rate is 1 frame per second (fps). We have $L_i = \lceil T_i/M \rceil$ chunks, where $\lceil \cdot \rceil$ denotes the ceiling function.

We adopt the pretrained language-image model CLIP \cite{radford2021learning} to extract per frame visual token features. For the $j^{th}$ video chunk $V_i^j \in \mathbb{R}^{M\times H \times W \times 3}$, we obtain $M\times h \times w$ visual tokens $F_{i}^{j}\in \mathbb{R}^{M\times h \times w \times D}$ from the CLIP vision encoder, where $h=H/p, w=W/p$, $D$ denotes the number of channels. $p$ denotes the patch size ($p=14$ for ViT-L/14, $h=16$, $w=16$).

\textbf{Chunk wise compression} The preliminary token number of a chunk is large, which would result in a high computational burden and large memory requirement for the LLM even though we only select a few chunks as input to LLM. Therefore, we perform chunk-wise pooling to obtain reduced spatial and temporal tokens (see Fig.~\ref{fig:STPooling} for the illustration in Supplementary Section \ref{sec:chunk}).

Particularly, we found that reducing the spatial resolution by a factor of 4 leads to marginal difference in performance while this can significantly reduce the number of tokens by 75\%. Therefore, we perform spatial average pooling with stride 2 to have $M$ $\times$ $\bar{h}$ $\times$ $\bar{w}$  tokens per chunk, where $\bar{h}$=$h/2$ and $\bar{w}$=$w/2$. This is equivalent to taking
the CLIP features of reduced resolution as the extracted feature. 

How to further reduce the number of tokens while preserving spatial-temporal features of a chunk? Motivated by Video-ChatGPT \cite{maaz2023video},
we perform spatial-temporal pooling on the spatio-temporal features. Particularly, global spatial average pooling for each frame is performed and thus we obtain $M$ tokens for the $M$ frames (\egno, 4 tokens). Temporal pooling for each spatial position is performed to have $N=\bar{h} \times \bar{w} = 8 \times 8 = 64$
tokens. We have $N+M=64+4=68$ tokens $\bar{F}_i^j \in \mathbb{R}^{(N+M)\times D}$, with $D$ dimension for each token. Please see Supplementary for more details. Compared with the original 1024 in a chunk, the number of tokens has been reduced to 6.6\%. We store the reduced chunk tokens in a memory to facilitate retrieval for the later LLM inference.



In contrast to Video-ChatGPT \cite{maaz2023video},
which performs global spatial and temporal pooling over the entire video, ours with chunks is capable of \emph{preserving local details} and is suited for long video reasoning.

\subsection{Question-guided Retrieval for Chunk Selection}

For a long video $V_i$ with abundant video chunks, we retrieve and select the $K$ most relevant chunks based on the question/query and use them as input to the LLM. 
The top $K$ retrieval aims to efficiently identify the most informative video segments for answering the given question, reducing the memory and computational burden to LLM and excluding the interference from irrelevant content. 

Note that achieving this is non-trivial, as we do not have any ground-truth locations of the question relevant chunks for supervision. We propose learnable retrieval, i.e., the learnable selection of question-related video chunks, through end-to-end training. Particular, we add an MLP block to the text features to facilitate the matching. Another alternative design is to apply an MLP to the vision features during retrieval or apply MLPs to both the text encoder and the visual encoder.  For a video question pair in inference, if we apply the MLP to the text encoder, we only need the inference through the MLP once. If we apply the MLP to the vision encoder, we need the inference through the MLP for every chunk. For simplicity, we apply to the text feature.  

We encode the question $Q_i$ into a feature vector $\mathbf{q}_i \in \mathbb{R}^{D}$ using the frozen CLIP text encoder $f_{\theta}$ and a learnable MLP block $\psi$ as  
\begin{align}
\mathbf{q}_i = \psi(f_{\theta} (Q_i)),
    \label{eq: query}
\end{align}
where the MLP block is a two-layer perceptron consisting of two fully connected layers and an ReLU activation function in between. This MLP transformation strengthens the correlation between the question representation and the potential corresponding chunk features for chunk selection. The MLP block is very lightweight with negligible number of parameters (\ieno, 1.8M), while facilitating the question-aware retrieval for long video understanding.

To identify the question-related chunks, we measure the affinity between the question and the chunks.  
We define the similarity score between the question representation $\mathbf{q}_i$ and the $j^{th}$ chunk of video $V_i$ as the average cosine similarity between $\mathbf{q}_i$ and the $N+M$ chunk tokens as
\begin{align}
s_i^j = \frac{1}{N+M}\sum_{n=1}^{N+M} \frac{\mathbf{q}_i \cdot \mathbf{u}_{i,n}^j}{\lVert \mathbf{q}_i \rVert \lVert \mathbf{u}_{i,n}^j \rVert},
\label{eq: similarity}
\end{align}
where $\mathbf{u}_{i,n}^j$ denotes the $n^{th}$ token in the $j^{th}$ chunk (of video $V_i$), $n=1,\ldots, N+M$.
We rank the video chunks based on their similarity scores $s_i^j$ (where $j=1, \ldots, L_i$) and select the top $K$ most relevant chunks. The $K\times (N+M)$ visual tokens of these chunks are input to the LLM after a linear projection (a fully connected layer) on each token. 

\subsection{End-to-End Optimization}
We train the network using an end-to-end optimization approach with the video instruction data. The image encoder, text encoder, and the LLM are frozen during the training. Only the parameters of the MLP block and projector are optimized. For a video-text pair, we introduce soft matching (SM) loss $\mathcal{L}_{i}^{\text{SM}}$ to regularize the similarity learning as
\begin{align}  
\mathcal{L}_{i}^{\text{SM}} = - \frac{\mathbf{q}_i \cdot \bar{\mathbf{v}}_i}{\lVert \mathbf{q}_i \rVert \lVert \bar{\mathbf{v}}_i \rVert}, \quad \text{where} \: \bar{\mathbf{v}}_i = \frac{\sum_{j=1}^{L_i} e^{s_i^j} \mathbf{v}_i^j}{\sum_{j=1}^{L_i} e^{s_i^j}}, 
\label{eq: L_SM}  
\end{align} 
where $\bar{\mathbf{v}}_i$ is a weighted combination of the $\mathcal{L}_i$ chunk features.
For a weight $e^{s_i^j}$, $e$ denotes the base of the natural logarithm, $s_i^j$ denotes the similarity score between the query and the $j^{th}$ chunk features (see Eq. (2)) of the video $V_i$. $\mathbf{v}_i^j$ denotes the averaged token feature of the $j^{th}$ chunk, \ieno, $\mathbf{v}_i^j = \frac{1}{N+M} \sum_{n=1}^{N+M}\mathbf{u}_{i,n}^j$. The soft matching loss aims to align the aggregated visual feature with the question feature by adjusting weights. Through the optimization of the MLP block, high weights are assigned to video chunks closely related to the question and low weights to less relevant chunks, thus improving the selection of relevant chunks for accurate question answering. 

The overall loss is 
\begin{align}  
\mathcal{L}_i = \mathcal{L}^{pred}_i + \lambda \mathcal{L}_{i}^{\text{SM}}, 
\label{eq: loss}  
\end{align}  
where $\mathcal{L}^{pred}_i$ denotes the cross-entropy loss between the generated prediction and groundtruth answer (a.k.a. LLM loss). $\lambda$ is a hyper-parameter (we set to 10) that balances the contribution of the regularization term (see Table~\ref{tab:lambda} for the ablation).

\section{Experiments}
\label{Experiments}

\subsection{Implementation Details}
\label{subsec:implementation}
Following Video-ChatGPT \cite{maaz2023video}, we use LLaVA-7B \cite{liu2023visual} with CLIP ViT-L/14 \cite{radford2021learning} as both image and text encoders. The MLP block has two FC layers of 1024 neurons, and the projector has 4096 neurons. We fine-tune only the MLP and projector, keeping encoders and LLM frozen. With $K=5$, we obtain 340 visual tokens, comparable to Video-ChatGPT’s 356.
Unlike Video-ChatGPT, which trains only a 4.2M-parameter projector, we also train a 1.8M-parameter MLP, totaling 6.0M trainable parameters. Please see Supplementary Section \ref{subsec:more_implementation} for more details.



\subsection{Datasets and Evaluation}

\begin{table}[t]
\centering
\caption{Information of the evaluation datasets, including the number of videos, QA pairs, and the average video duration.}
\footnotesize
\resizebox{0.8\linewidth}{!}{
    \begin{tabular}{cccc}
    \hline
    {Dataset} & {\#Video} & {\#QA} & {Duration (sec.)} \\
    \hline
    Activitynet-QA & 800   & 8000  & 114 \\
    EgoSchema & 5031  & 5031  & 180 \\
    Next-QA & 570   & 4996  & 44 \\
    IntentQA & 567   & 2134  & 47 \\
    WildQA & 261   & 652   & 71 \\
    QaEgo4D & 166   & 1854  & 495 \\
    lifeQA & 49    & 372   & 74 \\
    Social-IQ 2.0 & 144   & 876   & 60 \\
    \hline
    \end{tabular}%
    }
\label{table:eval datasets}
\end{table} 

We use Video Instruction Data collected by Maaz  \etal \cite{maaz2023video}
for video instruction tuning. This dataset contains about 100k question and answer pairs based on the Activitynet-QA dataset (average duration 180 seconds).

We evaluate the generalization (zero shot) performance of our framework on long video QA datasets Activitynet-QA \cite{caba2015activitynet}, EgoSchema \cite{mangalam2024egoschema}, NExt-QA\cite{xiao2021next}, and IntentQA \cite{li2023intentqa}. To widely evaluate the effectiveness of our model, besides the four popular datasets, we also tested on other long video QA datasets WildQA \cite{castro-etal-2022-in-the-wild}, QaEgo4D \cite{Baermann_2022_CVPR}, lifeQA \cite{castro-etal-2020-lifeqa}, and Social-IQ 2.0 \cite{siq2}.  Table \ref{table:eval datasets} shows specific information about each dataset. We perform ablation studies on Activitynet-QA, WildQA, QaEgo4D, lifeQA, and Social-IQ 2.0. 

We evaluate performance using accuracy and a 0–5 similarity score following Video-ChatGPT \cite{maaz2023video} (see Supplementary Section \ref{sec:metrics} for details).
\begin{table}[t]
  \centering
  \caption{Performance comparison with the state-of-the-art methods for video QA on Activitynet-QA.}
  \resizebox{0.64\linewidth}{!}{
    \begin{tabular}{cc}
    \hline
    Model & Acc./Score \bigstrut \\
    \hline
    Video-LLaMA \cite{zhang2023video} & 12.4/1.1 \bigstrut\\
    FrozenBiLM \cite{yang2022zero} & 24.7/- \\
    VideoChat \cite{li2023videochat} & 26.5/2.2 \\
    LLaMA-Adapter \cite{gao2023llama} & 34.2/2.7 \\
    Elysium \cite{wang2024elysium} & 43.4/2.9 \\
    Video-LLaVA \cite{lin2023video} & 45.3/3.3 \\
    MovieChat \cite{song2024moviechat} & 45.7/3.1 \\
    BT-Adapter \cite{liu2024bt} & 46.1/3.2 \\
    Chat-UniVi \cite{jin2024chat} & 46.1/3.3 \\
    MiniGPT4-video(7B) \cite{ataallah2024minigpt4} & 46.3/- \\
    LLaMA-VID(7B) \cite{li2023llama} & 47.4/3.3 \\
    Video-ChatGPT \cite{maaz2023video} & 47.8/3.3 \\
    VideoChat2 \cite{li2024mvbench} & 49.1/3.3 \bigstrut\\
    \hline
    R-VLM (Ours) & \textbf{49.7/3.3} \bigstrut\\
    \hline
    \end{tabular}%
    }
  \label{tab:SOTA-Activitynet}%
  \vspace{-2mm}
\end{table}%


\begin{table}[t]
  \centering
  \caption{Performance comparison with the state-of-the-art methods for video QA on EgoSchema fullset.}
  \resizebox{0.47\linewidth}{!}{
    \begin{tabular}{cc}
    \hline
    Model & Acc. \bigstrut\\
    \hline
    SeViLA \cite{yu2024self} & 22.7 \\
    FrozenBiLM \cite{yang2022zero} & 26.9 \\
    mPLUG-Owl \cite{ye2023mplug} & 31.1 \\
    Intervideo \cite{wang2022internvideo} & 32.1 \\
    TimeChat \cite{ren2024timechat} & 33.0 \\
    LLoVi \cite{zhang2023simple} & 33.5 \\
    Vamos(13B) \cite{wang2023vamos} & 36.7 \\
    \hline
    R-VLM (Ours) & \textbf{40.2} \bigstrut\\
    \hline
    \end{tabular}%
    }
  \label{tab:SOTA-EgoSchema_fullset}%
\end{table}%


\begin{table}[t]
  \centering
  \caption{Performance comparison with the state-of-the-art methods for video QA on NExt-QA.\protect\footnotemark}
  \resizebox{0.78\linewidth}{!}{
    \begin{tabular}{ccccc}
    \hline
    Model & Des. & Tem. & Cau. & Acc. \\
    \hline
    Just-Ask \cite{yang2021just} & 36.0 & 31.8 & 30.4 & 38.4 \\
    CLIP \cite{radford2021learning} & 57.0 & 38.1 & 43.6 & 43.9 \\
    InternVideo \cite{wang2022internvideo} & \textbf{65.1} & 43.4 & 48.0 & 49.1  \\
    Mistral(7B) \cite{jiang2023mistral} & - & - & - & 51.1 \\
    VFC \cite{momeni2023verbs} & 64.1 & 45.4 & 51.6 & 51.5 \\
    LLoVi(7B) \cite{zhang2023simple} & 63.2 & \textbf{55.6} & 47.9 & 54.3 \\
    MVU(13B) \cite{ranasinghe2024understanding} & 64.1 & 55.4 & 48.1 & 55.2 \\
    Video-ChatGPT \cite{maaz2023video} & 62.7 & 48.8 & 58.8 & 56.2 \\
    \hline
    R-VLM (Ours) & 63.1 & 52.0 & \textbf{59.9} & \textbf{58.7} \\
    \hline
    \end{tabular}%
    }
  \label{tab:SOTA-NExt}%
  \vspace{-2mm}
\end{table}%

\begin{table}[t]
  \centering
    \centering
    \caption{Performance comparison with the state-of-the-art methods for video QA on IntentQA dataset.}
    \resizebox{0.41\linewidth}{!}{
      \begin{tabular}{cc}
      \hline
      Model & Acc. \\
      \hline
      Random & 20.0  \\
      HGQA \cite{xiao2022video} & 47.7 \\
      VGT \cite{xiao2022videographtransformervideo} & 51.3 \\
      BlindGPT \cite{ouyang2022training} & 51.6 \\
      Mistral(7B) \cite{jiang2023mistral} & 50.4 \\
      LLoVi(7B) \cite{zhang2023simple} & 53.6 \\
      CaVIR \cite{li2023intentqa} & 57.6 \\
      \hline
      R-VLM (Ours) & \textbf{57.7} \\
      \hline
      \end{tabular}%
    }
    \label{tab:SOTA-intentqa}
\end{table}%

  \vspace{3mm} 
  
\begin{table}[t]
  \centering
    \centering
    \caption{Comparison with the other video-language models, including Video-LLaMA \cite{damonlpsg2023videollama} and the baseline method Video-ChatGPT \cite{maaz2023video} (accuracy~(\%)/ average score).}
    \resizebox{1\columnwidth}{!}{
      \begin{tabular}{cccc}
       \hline
      \multicolumn{1}{c}{Dataset} & Video-LLaMA  & Video-ChatGPT  & R-VLM (Ours) \\
      \hline
      WildQA & 63.19/3.18 & 58.00/3.30 & \textbf{64.82/3.39} \\
      QaEgo4D & \textbf{35.35}/1.94 & 29.74/2.43 & 32.51/\textbf{2.45} \\
      lifeQA & 35.75/2.32 & 33.87/2.55 & \textbf{38.71/2.61} \\
      Social-IQ 2.0 & 55.78/2.90 & 57.73/3.26 & \textbf{63.65/3.40} \\
      Average & 47.52/2.59 & 44.84/2.89 & \textbf{49.92/2.96} \\
      \hline
      \end{tabular}%
    }
    \label{tab:SOTA}
    \vspace{-3mm}
\end{table}

\subsection{Comparison with Other Models}

We compare our final scheme \emph{R-VLM} (Retrieval based  video-language model) with the baseline method \emph{Video-ChatGPT} \cite{maaz2023video} and the other state-of-the-art methods. We treat \emph{Video-ChatGPT} as our baseline method, where we use the same training dataset, LLM of LLaMA-7B, and visual encoder CLIP-L as \emph{Video-ChatGPT}. 

Table \ref{tab:SOTA-Activitynet}, \ref{tab:SOTA-EgoSchema_fullset}, \ref{tab:SOTA-NExt}, and Table \ref{tab:SOTA-intentqa} show the performance comparison on Activitynet-QA, EgoSchema, NExt-QA, and IntentQA, respectively. Note that EgoSchema, NExt-QA, and IntentQA were not involved in training but are only used for testing. Our \ours{} achieves the state-of-the-art performance. On Activitynet-QA, our \ours{} outperforms LLaMA-VID-7B by 2.3\% in accuracy. Our \ours{} achieves competitive performance with VideoChat2, even though VideoChat2 uses much more training data than ours (1.1 million vs. our 100K video-text pairs). 

On WildQA, QaEgo4D, lifeQA, and Social-IQ 2.0, Table \ref{tab:SOTA} shows the comparison with the baseline method \emph{Video-ChatGPT}, and the method of \emph{Video-LLaMA} \cite{damonlpsg2023videollama}.  \ours{} outperforms \emph{Video-ChatGPT} significantly by \textbf{6.8}\%, \textbf{2.8}\%, \textbf{4.8}\%, \textbf{6.0}\% in accuracy, respectively. Ours consistently achieves the best average score. Note that the number of visual tokens of our \ours{} is comparable to that of \emph{Video-ChatGPT} (\ieno, 340 vs. 356), making a fair comparison. This demonstrates the effectiveness of our learnable retrieval-based design, which facilitates the exploration of the most informative visual tokens and preservation of necessary vision details for QA. We found that when evaluating Video-LLaMA, the metric of accuracy is not as reliable as the score. \emph{Video-LLaMA} is prone to give detailed description of the entire video, whereas the answer is usually not question-specific. The accuracy metric treats such answer as correct, while the score rating is relatively reasonable. Fig.~\ref{fig:good_qaego4d_3} in Supplementary Section \ref{sec:morevis} shows a typical example on QaEgo4D, where the answer includes much irrelevant (or incorrect) information while the answer is submerged in the overall answer. In contrast, our model provides more concise and accurate answers. This explains why our accuracy is lower than Video-LLaMA on QaEgo4D.

\begin{table*}[thbp]
\centering
\caption{Comparison (accuracy (\%) / average score) of different methods and ablation studies. All these models are trained using the same video instruction data. \emph{R-VLM} denotes our final scheme with learnable retrieval. \emph{Video-ChatGPT} is our baseline method. \emph{R-VLM w/ Uni.} denotes uniform sampling of $K$ chunks in our framework instead of retrieval-based sampling.  \emph{R-VLM w/ CLIP M.} denotes that we use the final CLIP class token features of vision and text for matching without incorporating learnable parameters for retrieval. \emph{R-VLM w/o SM} denotes that we do not use the proposed soft matching (SM) loss. \emph{R-VLM w/o G.} denotes that the gradient back-propagation from the LLM loss $\mathcal{L}^{pred}_i$ to the learnable MLP block $\psi$ is disabled. }
\resizebox{0.98\linewidth}{!}{
    \begin{tabular}{c|c|ccccc}
    \hline
    Dataset & Video-ChatGPT & R-VLM w/ Uni. & R-VLM w/ CLIP M. & R-VLM w/o SM & R-VLM w/o G. & R-VLM \\
    \hline
    ActivityNet-QA & 47.81/3.27 & 49.34/3.32 & 48.86/3.30 & 48.39/3.32 & 48.30/3.31 & \textbf{49.65/3.34} \\
    WildQA & 58.00/3.30 & 61.23/3.36 & 60.31/3.27 & 59.94/3.28 & 62.27/3.32 & \textbf{64.82/3.39} \\
    QaEgo4D & 29.74/2.43 & 31.57/2.44 & 31.52/2.43 & 31.12/2.36 & 31.66/2.46 & \textbf{32.51/2.45} \\
    lifeQA & 33.87/2.55 & 36.56/2.56 & 31.45/2.42 & 36.29/2.47 & 37.09/2.60 & \textbf{38.71/2.61} \\
    Social-IQ 2.0 & 57.73/3.26 & 57.96/3.24 &  61.17/3.28  & 57.22/3.17 & 62.43/3.36 & \textbf{63.65/3.40} \\
    \hline
    Average & 45.43/2.96 & 47.33/2.98 & 46.66/2.94 & 46.59/2.92 & 48.35/3.01 & \textbf{49.87/3.03} \\
    \hline
    \end{tabular}
    }%
\label{table: comparison}
\end{table*}

\subsection{Complexity Analysis}

Section \ref{subsec:complexity} in Supplementary provides detailed analysis.

\subsection{Ablation Studies}

We study the effectiveness of our retrieval designs, the soft matching loss, chunk-wise design, the influence of hyperparameter $K$, comparison with Q-Former-based designs, etc.

\vspace{1mm}

\noindent\textbf{Effectiveness of Retrieval for Chunk Selection}
We compare our retrieval mechanism and the uniform sampling strategy for selecting $K=5$
chunks as LLM input. In the uniform sampling setting (\emph{R-VLM w/ Uni.}), we select $K$ chunks uniformly rather than by question guided retrieval. Table ~\ref{table: comparison} shows our scheme \ours{} with learnable retrieval outperforms \emph{R-VLM w/ Uni.} by 3.6\%, 0.9\%, 2.2\%, and 5.7\% on WildQA, QaEgo4D, lifeQA, and Social-IQ 2.0, respectively. 

Since question-relevant chunks usually form only a small portion of a long video, uniform sampling often misses them, leading the LLM to wrong answers. Our learnable retrieval instead selects the most relevant chunks, providing reliable, non-redundant context. Visualizations in Fig.~\ref{fig:good_qaego4d_2} further show its interpretability by indicating which chunks support the answer.


\vspace{2mm}

\noindent\textbf{Effectiveness of Learnable Retrieval vs. Off-the-shelf CLIP based Retrieval} 
We introduce learnable retrieval via the MLP block. To compare, we design \emph{R-VLM w/ CLIP M.} that which uses the averaged class token feature of the last layer of the CLIP image encoder as the vision chunk feature, and the class token feature of the last layer of the CLIP text encoder as the question feature for matching, with no learnable parameters.
Table~\ref{table: comparison} shows our \ours{} with learnable retrieval consistently outperforms \emph{R-VLM w/ CLIP M.}, up to 7.3\% on LifeQA. This is likely because CLIP is trained for image-caption matching, whereas our MLP-adapted retrieval better aligns question and relevant video chunks.

\noindent\textbf{Influence of Soft Matching (SM) Loss} SM loss regularizes MLP learning for better retrieval. Table~\ref{table: comparison} shows our framework with SM loss (\emph{R-VLM}) outperforms the version without it (\emph{R-VLM w/o SM}).
SM loss re-weights video chunk tokens based on cosine similarity with the text embedding, then maximizes similarity between the video chunk tokens and the text embedding. This facilitates our learnable retrieval layer to identify the video clips most related to the questions.


\vspace{2mm}

\noindent\textbf{Influence of LLM Loss to Retrieval} The instruction tuning with the LLM loss ($\mathcal{L}^{pred}_i$) also propagates the gradient to the learnable MLP block $\psi$,  which facilitates the selection of right chunks for retrieval. When we disable the back-propagation to $\psi$, we have scheme \emph{R-VLM w/o G.}. Table~\ref{table: comparison} shows our final scheme \emph{R-VLM} that allows the back-propagation obviously outperforms \emph{R-VLM w/o G.} by 2.6\%, 0.9\%, 1.6\%, and 1.2\% on WildQA, QaEgo4D, lifeQA, and Social-IQ 2.0.


\vspace{2mm}

\noindent\textbf{Effectiveness of Chunk-wise Design} \emph{Video-ChatGPT} performs video level spatial temporal pooling to obtain 356 visual tokens. Such global pooling would result in loss of details, especially when the
question-related
video segments take a small portion of the entire video. In our design, we perform chunk level spatial temporal pooling to preserve more information about each chunk. Table \ref{table: comparison} shows that when we
uniformly
sample the chunks to have 340 visual tokens, \emph{R-VLM w/Uni} obviously outperforms \emph{Video-ChatGPT}, demonstrating the effectiveness of our chunk-wise design. 

\vspace{2mm}

\noindent\textbf{Influence of Hyperparameter $K$} A small $K$ may miss necessary information, while a large $K$ may introduce interference. Table \ref{table: ablation K} shows that as $K$ gradually increases from 1 to 5, the average performance increases. When $K$ increases from 5 to 7, the performance decreases. We found $K=5$ presents a good trade-off on most datasets, even though there are slight differences on different datasets. 
In addition, we calculated the computation (FLOPs) cost saving (measured by percentage) using different $K$, compared with that of using all chunks (see Section \ref{subsec:complexity} in Supplementary). Table \ref{table: ablation K} shows that given $K=5$, the percentage of saving ranges from 64\% to 95\% for different datasets (due to different video lengths), demonstrating the high efficiency of our solution. For ActivityNet-QA, the percentage of saving ranges from 93\% to 73\% when $K$ is set to 1 to 7.
We leave the adaptive design of $K$ as future work.

\begin{table*}[t]
\centering
\renewcommand\tabcolsep{3mm}
\caption{Ablation study on the influence of $K$, evaluated in terms of accuracy (\%) /score /computation-saving~(\%) in LLM inference. We use bold to mark the best performance and underline to mark the second-best performance.}
\resizebox{0.7\linewidth}{!}{
\begin{tabular}{c|c|cccc}
 \hline 
Dataset & Video-ChatGPT & Ours(K=1) & Ours(K=3) & Ours(K=5) & Ours(K=7) \\
 \hline
ActivityNet-QA & 47.81/3.27 & 46.75/3.27/93 & 48.10/3.31/86 & \underline{49.65}/\textbf{3.34}/80 & \textbf{49.76}/\textbf{3.34}/73 \\
WildQA & 58.00/3.30 & 57.45/3.18/89 & 60.58/3.31/79 & \textbf{64.82}/\textbf{3.39}/69 & \underline{63.44}/\textbf{3.39}/59 \\
QaEgo4D & 29.74/\underline{2.43} & 32.42/2.41/98 & 32.04/2.42/97 & \underline{32.51}/\textbf{2.45}/95 & \textbf{32.81}/2.42/93 \\
lifeQA & 33.87/2.55 & 37.63/\underline{2.62}/90 & \underline{38.44/2.62}/80 & \textbf{38.71}/2.61/71 & 37.63/\textbf{2.65}/61 \\
Social-IQ 2.0 & 57.73/3.26 & \textbf{63.92/3.44}/87 & 60.89/3.34/76 & \underline{63.65/3.40}/64 & 62.89/3.34/52 \\
\hline
Average & 44.84/2.89 & 47.86/2.91/91 & 47.99/2.92/84 & \textbf{49.92/2.96}/76 & \underline{49.19/2.95}/68 \\
 \hline
\end{tabular}%
}
\label{table: ablation K}
\vspace{-2mm}
\end{table*}

\vspace{2mm}

\noindent\textbf{Comparison with Q-Former Designs} We compare our retrieval-based design with Q-Former variants. 1) \emph{VLM-QFormer} aggregates all chunk tokens ($L_i \times (N+M)$, $N+M=68$) into 340 visual tokens without retrieval. Table~\ref{tab:Qformer} shows \ours{} significantly outperforms \emph{VLM-QFormer}, thanks to the retrieval-based design which can exclude the inference of irrelevant visual tokens. 2) \emph{VLM-InstructQFormer} adds question-aware aggregation, similar to InstructBLIP \cite{dai2305instructblip}, improving over \emph{VLM-QFormer} but still underperforming \ours{}. This highlights the effectiveness of our retrieval-based design for extracting question-relevant visual features.

\begin{table}[t]
  \centering
  \caption{Comparison with Q-Former based two designs under our framework, evaluated in terms of accuracy (\%)/score.}
  \resizebox{1\linewidth}{!}{
    \begin{tabular}{cccc}
    \hline
     & VLM-Qformer & VLM-InstructQformer & R-VLM (Ours) \\
    \hline
    ActivityNet-QA & 49.34/\textbf{3.35} & 48.94/3.33 & \textbf{49.65}/3.34 \\
    WildQA & 58.59/3.16 & 58.28/3.17 & \textbf{64.82/3.39} \\
    QaEgo4D  & 30.58/2.35 & 31.23/2.41 & \textbf{32.51/2.45} \\
    lifeQA & 35.48/2.56 & 37.10/\textbf{2.62} & \textbf{38.71}/2.61 \\
    Social-IQ 2.0 & 54.30/3.14 & 57.62/3.19 & \textbf{63.65/3.40} \\
    Average & 45.66/2.91 & 46.64/2.94 & \textbf{49.87/3.03} \\
    \hline
    \end{tabular}%
    }
  \label{tab:Qformer}%
  \vspace{-4mm}
\end{table}%

\noindent\textbf{Influence of the Time Duration of a Chunk} Please see Section \ref{sec:more-ablation} in Supplementary for detailed analysis.

\noindent\textbf{Influence of MLP Block Designs} For the MLP block, we use two-layer MLP by default. We found that using deeper layers achieves similar performance (see Supplementary VII.C).

\subsection{Visualization Analysis}

Section \ref{sec:morevis} in Supplementary shows visualization.

\section{Conclusion}

The comprehension of long videos using LLMs remains an under-explored area. There are two main challenges associated with comprehending long videos. 1) Long videos generally lead to abundant visual tokens, which increase  computational cost for LLM inference. 2) Global aggregation of visual tokens inevitably results in the loss of vision details especially when the question relevant video chunks take only a small portion of the entire video. Moreover, question irrelevant chunks introduce interference. In this work, we address these issues by introducing a simple yet effective retrieval-based video-language model (R-VLM) for long-video understanding. 
Our R-VLM effectively reduces the number of video tokens, preserves the most informative information, eliminates noise interference, and thus enhances system performance. Our experimental results demonstrate the effectiveness of our designs for comprehending long videos.


\bibliography{tmm}
\bibliographystyle{IEEEtran}

\clearpage
\twocolumn[
\begin{center}
    {\Huge {Supplementary Material}}
\end{center}
]

\section{More Details on Visual Token Extraction}
\label{sec:chunk}
\begin{figure*}[th]
	\centering
	\includegraphics[width=0.5\textwidth]{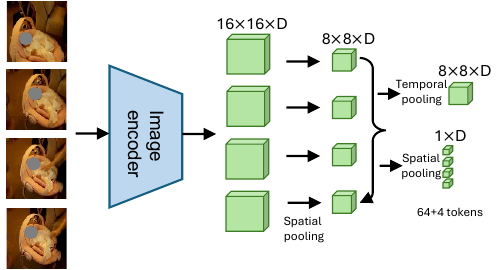}
    \caption{Illustration of the spatial and temporal pooling to obtain 68 visual tokens for a chunk. We perform spatial average pooling with stride 2 to have $M$ $\times$ $\bar{h}$ $\times$ $\bar{w}$ = $4\times 8 \times 8 = 256$ tokens per chunk, where $\bar{h}$=$h/2$ and $\bar{w}$=$w/2$. This is equivalent to taking the CLIP features of reduced resolution as the extracted feature. Global spatial average pooling for each frame is performed and thus we obtain $M$ tokens for the $M$ frames ($M=4$). Temporal pooling for each spatial position is performed to have
$N=\bar{h} \times \bar{w} = 8 \times 8 = 64$
tokens. Therefore, we have $N+M=64+4=68$ tokens for a chunk.}
    \vspace{-5mm}
    \label{fig:STPooling}
\end{figure*}

To better illustrate the procedure of visual token extraction of a chunk, Fig.~\ref{fig:STPooling} illustrates the spatial and temporal pooling for obtain 68 visual tokens for a chunk.

\section{More Experiments}

\subsection{More Implementation Details}
\label{subsec:more_implementation}

Following Video-ChatGPT \cite{maaz2023video}, we use LLaVA 7B \cite{liu2023visual} as our base model. We utilize the pre-trained CLIP ViT-L/14 \cite{radford2021learning} as our image encoder and extract the feature from the second-to-last layer as the $h \times w$ visual tokens of a frame. We use the fine-tuned Vicuna (7B) from LLaVA as our LLM. For the text encoder, we use the pre-trained CLIP ViT-L/14 text encoder and extract the class token feature of the penultimate layer.
Each of the two fully connected layers within the MLP block is equipped with 1024 neurons.
The projector has 4096 neurons.

We only fine-tune the MLP block and the projector while keeping the image encoder, the text encoder and the LLM frozen. We fine-tune the model for 3 epochs using video instruction data, with a learning rate of 2e-5 and a batch size of 40. Training our model takes about 24h on an A100 80GB GPU.
We set $K$ to 5, resulting in $5\times (64+4) = 340$ visual tokens as the input to the LLM. This is comparable to the number of visual tokens used in Video-ChatGPT (356 visual tokens). 

For VideoChat-GPT, only the projector of 4.2M parameters is trained. In our framework, besides the projector, we added a small trainable MLP block of 1.8M parameters. Our total number of trainable parameters is 6.0M, which is very small considering the total model size is 6.7B.

\subsection{More Details on Evaluation Metrics}
\label{sec:metrics}

In this paper, we follow the metrics of accuracy and average score as proposed by Video-ChatGPT \cite{maaz2023video}
for performance evaluation, where we use ChatGPT (gpt-3.5-turbo) to assist in judging the correctness of model predictions. ChatGPT accepts questions, ground-truth answers, and the model predictions as input. For each question-answer
pair, ChatGPT gives a binary judgment of ``yes" or ``no" to identify whether the predicted answer is correct or not, for accuracy evaluation. Moreover, an integer score of 0-5 is also given by ChatGPT to indicate how similar the prediction is to the answer. 0 represents the lowest score and 5 represents the highest score.

\subsection{More Results on Other Datasets}
\label{sec:other-datasets}

We also report the results on the subset of EgoSchema in Table~\ref{tab:SOTA-EgoSchema_subset}. It can be seen that our R-VLM method achieves the best performance.

\begin{table}[t]
  \centering
  \caption{Performance comparison with the state-of-the-art methods for video QA on the EgoSchema subset of 500 testing pairs.}
  \resizebox{0.44\linewidth}{!}{
    \begin{tabular}{cc}
    \hline
    Model & Acc. \bigstrut\\
    \hline
    ViperGPT \cite{suris2023vipergpt} & 15.8  \\
    Video-ChatGPT \cite{maaz2023video} & 20.0 \\
    SeViLA \cite{yu2024self} & 25.7 \\
    mPLUG-Owl \cite{ye2023mplug} & 33.8 \\
    Video-LLaVA \cite{lin2023video} & 36.8 \\
    \hline
    R-VLM (Ours) & \textbf{46.0} \\
    \hline
    \end{tabular}%
    }
  \label{tab:SOTA-EgoSchema_subset}%
\end{table}%

\subsection{Computational Complexity}
\label{subsec:complexity}


\begin{table}[t]
  \centering
  \caption{Average time cost for 60 seconds videos from Social-IQ 2.0. The first part is feature extraction and the second part is retrieval and LLM inference.}
  \resizebox{1\linewidth}{!}{
    \begin{tabular}{cccc}
    \hline
    {} & Feature Extraction & Retrieval \& LLM inference  & Total \\
    \hline
    Time~(seconds) & 0.14  & 2.42 & 2.56 \\
    \hline
    \end{tabular}%
    }
  \label{tab:time_consummation}%
\end{table}%

The computational cost of our framework mainly comes from two parts. The first part is to encode the video frames through the CLIP encoder and the spatial-temporal pooling to get chunks. The second part is the retrieval of
$K=5$ chunks and put them to the LLM for inference.

On a single A100, we tested 120 videos with 60 seconds each from Social-IQ 2.0 and calculated the average inference time for a video, as shown in Table \ref{tab:time_consummation}. 
For a video, the first part for vision feature extraction takes an average of 0.14 seconds (in parallel for 60 frames), and the second part (mainly LLM inference) takes an average of 2.42 seconds. The total time is 2.56 seconds. The retrieval by encoding the question representation, and calculating the similarity scores between the question representation and the chunks, is very fast with almost negligible time. 
Note that for a longer video, the time consumption for LLM inference does not increase since the input number of visual tokens to the LLM is fixed (\ieno, 68$\times$5=340) in our scheme, which is favored for long video or streaming video understanding.

\begin{table}[t]
  \centering
  \caption{Analysis of FLOPs saving of our scheme for LLM inference. $Avg. chunks$ denotes the average number of video chunks. $S_{vis}$ denotes the average number of visual tokens of all the chunks of a video (obtained by multiplying $Avg. Chunks$ by the number of visual tokens per chunk (\ieno, 68)). $S_{vis}'$ denotes the average number of visual tokens of the selected $K=5$ chunks of a video (5 $\times$ 68 = 340).}
  \resizebox{1\linewidth}{!}{
    \begin{tabular}{cccccc}
    \hline
    {} & ActivityNet-QA & WildQA & QaEgo4D & lifeQA  & Social-IQ 2.0 \\
    \hline
    $Avg.~chunks$ & 29 & 19 & 122 & 20 & 16\\
    $S_{vis}$ & 1972 & 1292 & 8296 & 1360 & 1088\\
    $S_{vis}'$ & 340 & 340 & 340 & 340 & 340\\
    $Save$ & 80\% & 69\% & 95\% & 71\% & 64\%\\
    \hline
    \end{tabular}%
    }
  \label{tab:flops_saved}%
\end{table}%

The FLOPs for LLM inference can be roughly estimated as 2$PS$ \footnote{https://medium.com/@dzmitrybahdanau/the-flops-calculus-of-language-model-training-3b19c1f025e4}, where $P$ denotes the number of parameters, and $S$ denotes the number of tokens. The computational complexity of LLM is proportional to the number of tokens which consists of text tokens (question and answer) and visual tokens. The LLM model size $P$ is 6.7B. On the training dataset, the average number of tokens for question and answers is 80, \ieno, $S_{tex} = 80$. This varies on different testing datasets. For simplicity, we assume the number is the same for all the datasets. We denote the number of visual tokens as $S_{vis}$. The total number of tokens is then $S = S_{tex} + S_{vis}$. The saved computational cost (FLOPs) for LLM inference can be computed approximately as 
\begin{align}
 Saved=\frac{S_{vis} - S_{vis}'}{S_{tex} + S_{vis}}.
\label{eq: flops}
\end{align}

We show the computation saving of four long video datasets in Table \ref{eq: flops}. \textbf{Thanks to our retrieval, only $K=5$ chunks ($S_{vis}' = 5 \times 68 = 340$ tokens) instead of all the chunks are needed as the input to LLM. The saved computational cost (FLOPs) ranges from 69\% to 95\%}, demonstrating the high efficiency of our solution.

Moreover, our retrieval-based solution can significantly save the memory consumption. For an A100 GPU of 80G memory, given extracted chunk features, when we set batch size to 40 and $K$ to 5, the memory occupation is about 70G in training. In contrast, when a baseline scheme without using retrieval that all the chunks (instead of 5 selected chunks) are input to LLM, videos longer than 40 seconds are not able to be filled to the memory (out of memory) in training. In contrast, our scheme could support very long videos (even 80+ hours).

We have also estimated the memory cost for training different schemes when we set the batch size as 20 and use input video sequences of 80 seconds. Our scheme with $K$=5 uses similar size of memory (45.3 GB) as VideoChat-GPT (42.2 GB). In contrast, the total memory cost without using retrieval (input all the chunks, in total 1360 visual tokens) is 75.1GB, demonstrating that our retrieval-based design (45.3GB) can significantly save memory and is efficient.

\subsection{More Ablation Studies}
\label{sec:more-ablation}

\noindent\textbf{Influence of the Hyperparameter $\lambda$} We determine $\lambda$ based on our previous experiences, where we usually can obtain satisfactory results by setting different losses at the same magnitude. This is a simple way to ensure that different losses are considered in optimization. We found when we set $\lambda$ as 10, the two loss terms are at the similar magnitude. We also conducted ablation study on the choose of $\lambda$ by setting it to 1, 10, and 100. Table \ref{tab:lambda} shows the comparison. We can see that  $\lambda=10$ provides he best performance on average, slightly better than that of $\lambda=1$.

\begin{table}[t]
  \centering
  \caption{Ablation study on $\lambda$.}
  \resizebox{0.85\linewidth}{!}{
    \begin{tabular}{cccc}
    \toprule
    \textbf{} & \textbf{$\lambda$=1} & \textbf{$\lambda$=10} & \textbf{$\lambda$=100} \\
    \midrule
    ActivityNet-QA & \textbf{49.70}/3.32 & 49.65/\textbf{3.34} & 49.06/3.33 \\
    WildQA & 64.42/\textbf{3.41} & \textbf{64.82}/3.39 & 63.34/3.35 \\
    QaEgo4D  & \textbf{34.57}/\textbf{2.50} & 32.51/2.45 & 32.85/2.48 \\
    lifeQA & 36.49/2.56 & \textbf{38.71}/2.61 & 36.93/\textbf{2.67} \\
    Social-IQ 2.0 & 62.77/3.38 & \textbf{63.65}/\textbf{3.40} & 60.71/3.32 \\
    \hline
    Average & 49.59/3.03 & \textbf{49.87}/\textbf{3.04} & 48.58/3.03 \\
    \bottomrule
    \end{tabular}%
    }
  \label{tab:lambda}%
\end{table}%

\begin{table}[t]
  \centering
  \caption{Influence of the time duration of a chunk, evaluated in terms of accuracy (\%)/score. 8f/c denotes 8frames/chunk.}
  \resizebox{1\linewidth}{!}{
    \begin{tabular}{cccc}
    \hline
    \textbf{} & {Video-ChatGPT } & R-VLM(8f/c) & R-VLM(4f/c) \\
    \hline
    ActivityNet-QA & 47.81/3.27  & 49.24/3.31 & \textbf{49.65/3.34} \\
    WildQA & 58.00/3.30  & 62.58/3.37 & \textbf{64.82/3.39} \\
    QaEgo4D  & 29.74/2.43  & 32.36/2.43 & \textbf{32.51/2.45} \\
    lifeQA & 33.87/2.55 & 36.83/2.63 & \textbf{38.71/2.61} \\
    Social-IQ 2.0 & 57.73/3.26 & 62.20/3.33 & \textbf{63.65/3.40} \\
    \hline
    Average & 45.43/2.96 & 48.64/3.01 & \textbf{49.87/3.03} \\
    \hline
    \end{tabular}%
    }
  \label{tab:chunksize}%
\end{table}%

\noindent\textbf{Influence of the Time Duration of a Chunk} In considering the memory limitation and video redundancy, we sample at a low frame rate (1 fps). To allow a chunk to contain dynamic temporal information, we use 4 seconds that are expected to contain temporal dynamics as the chunk unit. We study the influence of the time duration of a chunk by comparing the use of 4 frames (4 seconds) and 8 frames (8 seconds) as a chunk at the same sample rate of 1 fps. Table \ref{tab:chunksize} shows the performance comparison. We can see that the performance of using 4 frames/chunk is better than that using 8 frames/chunk. 
Firstly, 4 frames/chunk allows identification of shorter temporal segments. Secondly, it preserves more spatial details than 8 seconds per chunk since an 8-second chunk pools features over a longer period. 
Both the two settings outperform the baseline method Video-ChatGPT.

\vspace{2mm}

\noindent\textbf{Influence of MLP Block Designs} For the MLP block, we use two-layer MLP on text encoder by default. As shown in Table \ref{tab:ablation_mlp}, we found that using deeper layers (six-layer MLP) achieves similar average performance.

As discussed in Section Section III.B, we add an MLP block after the text encoder to facilitate the matching. Another alternative design is to apply an MLP to the vision encoder for retrieval. We added an ablation comparison of adding MLP after vision encoder (R-VLM (2-layer MLP on vision)) vs. adding MLP after text encoder (R-VLM (2-layer MLP on text)). Table \ref{tab:ablation_mlp} shows that similar average performance is achieved.

\begin{table}[t]
  \centering
  \caption{Ablation study on MLP block designs: two layers vs. six layers after the text encoder, and MLP after the vision encoder. Performance is evaluated in terms of accuracy (\%)/score.}
  \resizebox{1\linewidth}{!}{
    \begin{tabular}{cccc}
    \hline
     & R-VLM & R-VLM & R-VLM  \\
     & (2-layer MLP & (6-layer MLP  & (2-layer MLP  \\
     & on text) & on text)  & on vision)  \\
    \hline
    ActivityNet-QA & 49.65/3.34 & 49.02/3.32 & 49.71/3.25 \\
    WildQA & 64.82/3.39 & 62.83/3.34 & 64.36/3.36  \\
    QaEgo4D  & 32.51/2.45 & 32.29/2.44 & 31.75/2.42 \\
    lifeQA & 38.71/2.61 & 42.30/2.65 & 39.33/2.59  \\
    Social-IQ 2.0 & 63.65/3.40 & 63.31/3.39 & 61.82/3.36 \\
    \hline
    Average & 49.87/3.03 & 49.95/3.03 & 49.39/3.00 \\
    \hline
    \end{tabular}%
    }
  \label{tab:ablation_mlp}%
\end{table}%

\begin{figure*}[!htbp]
    \centering
    \includegraphics[width=1.0\linewidth]{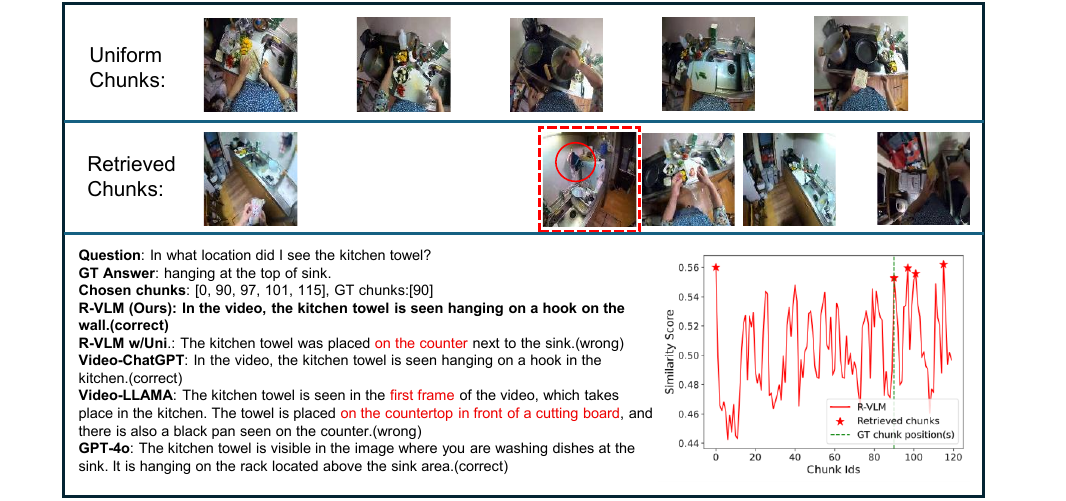}
    \caption{Visualization of video QA examples from QAEgo4D. The kitchen towel related to the question does not appear in the uniformly sampled video chunks. The second chunk selected by our model contains kitchen towel. Our answer states that the towel is hunging on a hook on the wall. Video-LLaMA answers incorrectly, where the towel does not appear in the first frame of the video, and it is not be placed on the countertop in front of a cutting board. Note that we use a sampled frame in a chunk to illustrate the chunk in the first two rows. GPT-4o correctly points out where the towel is.}
    \label{fig:good_qaego4d_2}


    \includegraphics[width=1.0\linewidth]{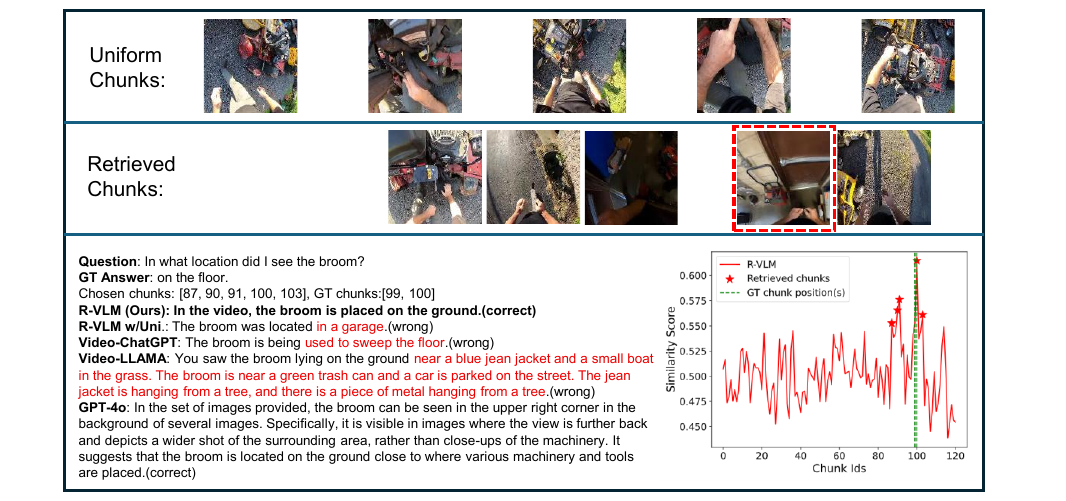}
    \caption{Visualization of a video QA example from QAEgo4D. The broom is small and is on the left in the red boxed image. Our R-VLM captures exactly where the broom is, i.e., on the ground. R-VLM w/Uni. does not capture the video chunks with broom and thus does not answer accurately. The answer of Video-ChatGPT is irrelevant to the question. The answer from Video-LLaMA is redundancy and tedious, where the mentioned blue jean jacket and boat actually do not appear in the video. GPT-4o gave the correct answer, \ieno, the broom is on the ground.}
    \label{fig:good_qaego4d_3}
\end{figure*}

\begin{figure*}[!htbp]
    \centering       \includegraphics[width=1.0\linewidth]{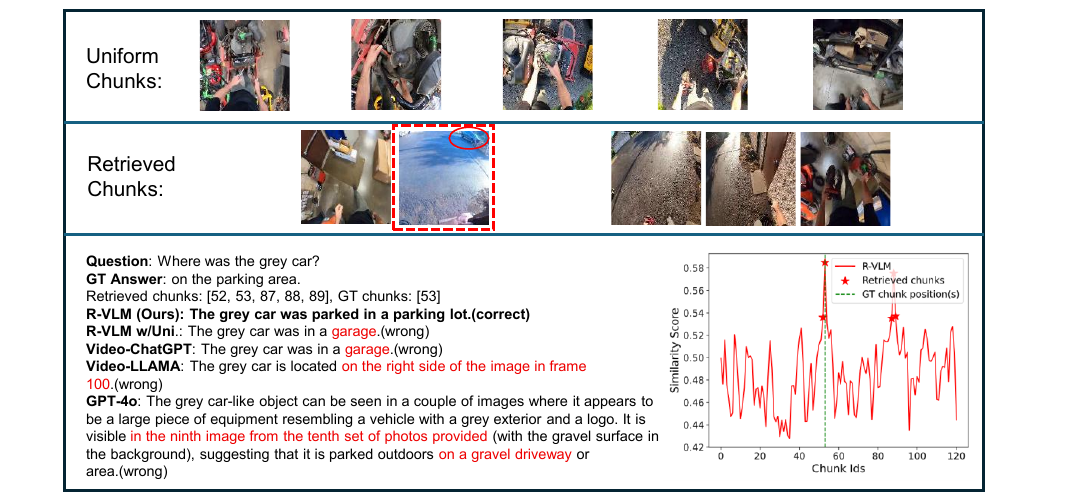}  
    \caption{Visualization of a video QA example from QAEgo4D. We can see that the gray car does not appear in the uniformly sampled video chunks. Our R-VLM correctly answers that the car was parked in the parking lot (outdoors), but R-VLM w/Uni.’s answer was the garage (indoors). Video-LLaMA does not answer where the car is and the ground-truth frames do not appear in the frame 100. Video-ChatGPT made the similar mistake as R-VLM w/Uni. GPT-4o had an illusion that the car was parked on a gravel road, but the road was flat, so it gave an inaccurate answer.}
    \label{fig:good_qaego4d_1}  
\end{figure*}

\begin{figure*}[!htbp]
    \centering   
    \includegraphics[width=1.0\linewidth]{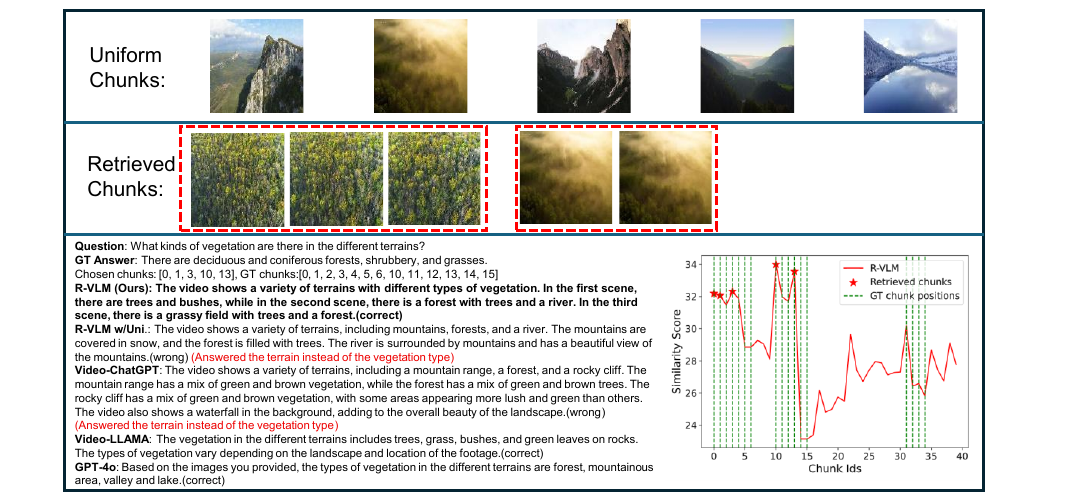} 
    \caption{Visualization of video QA examples from WildQA. In this video, two clips show vegetation and the remaining clips show mountains, rivers, etc. Uniform sampling mainly obtains segments such as mountains and rivers rather than segments with vegetation. Therefore, only the terrain was answered, without giving vegetation types. In contrast, our retrieved chunks contain video clips of vegetation. Thus the types of vegetation are predicted correctly: trees, bushes, forest. Video-ChatGPT gives a global description and does not answer specific vegetation types. GPT-4o gave correct answer.}      
    \label{fig:good_wildqa_3}  
\end{figure*}

\vspace{16mm}
\begin{figure*}[!htbp]  
    \centering  
        \centering  
        \includegraphics[width=\linewidth]{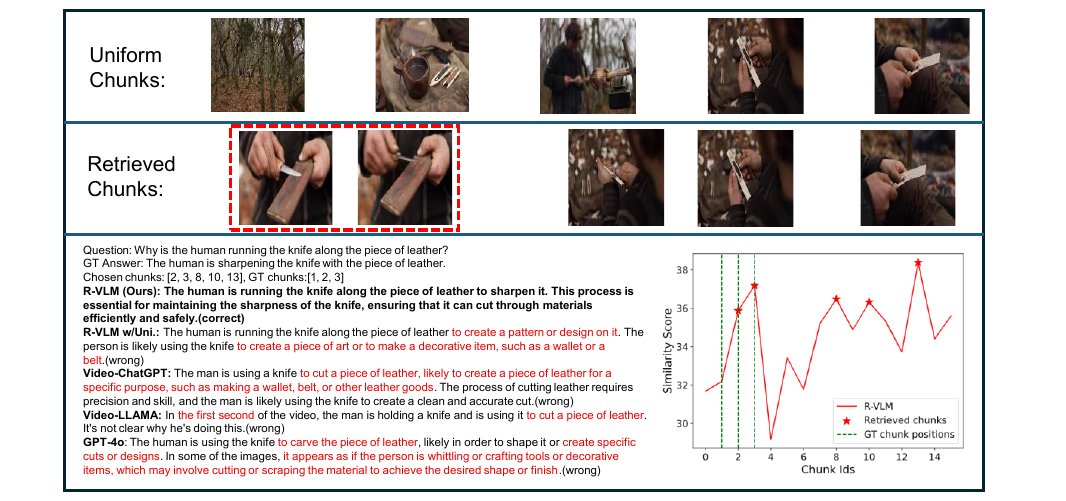}  
        \caption{Visualization of a video QA example from WildQA. The uniform sampling miss the chunks for sharpening process in ground-truth segments (at the beginning of video). As a result, LLM does not see the knife running along the leather, and only see the knife and some delicate small objects. Therefore, R-VLM w/Uni. mistakenly thought that this individual was carving patterns or making designs. Our retrieved chunks retain the process of the knife running on the leather and therefore R-VLM gives the correct answer. Both Video-LLaMA and Video-ChatGPT answered that people are cutting leather with a knife to make art, rather than sharpening the knife. GPT-4o thinks the knife is carving leather, which is inconsistent with the correct answer: sharpening the knife.}
        \label{fig:good_wildqa_1}  
\end{figure*}  

\begin{figure*}[!htbp]  
        \centering  
        \includegraphics[width=\linewidth]{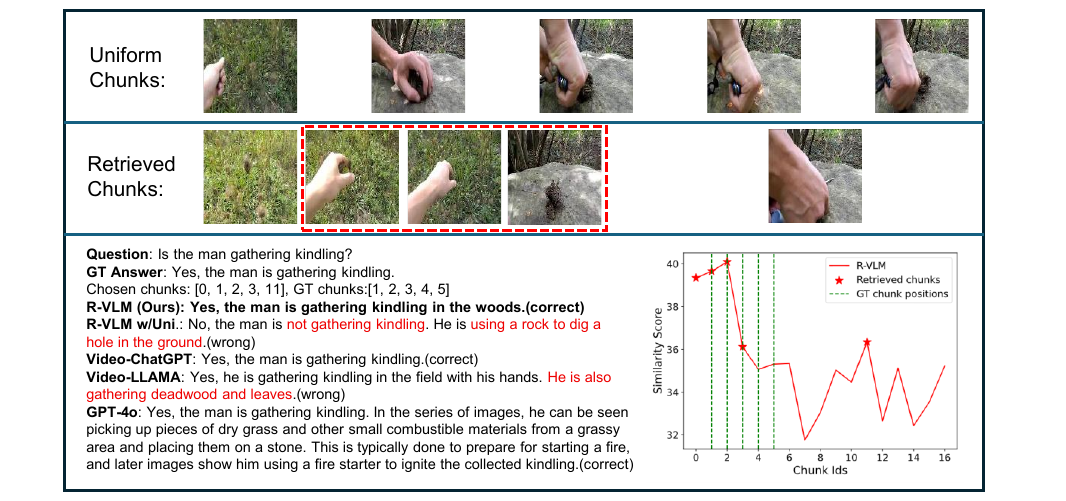} 
        \caption{Visualization of a video QA example from WildQA. In this video, collecting the kindling takes a short time, while placing the tinder on the stones takes a longer time. Uniform sampling makes LLM think that there is no process of collecting kindling and output the wrong answer of “digging a hole”. Our R-VLM identified the relevant chunks of “gathering” even though those chunks only take a small duration in the entire video, generating correct answer. Video-LLaMA’s prediction is not accurate since in fact the man did not gather deadwood and leaves in the video. GPT-4o gave the correct answer on gathering kindling.}     
        \label{fig:good_wildqa_2}   
\end{figure*}

\begin{figure*}[!htbp] 
    \centering  
    \begin{subfigure}{\mywidth\textwidth}  
        \centering  
        \includegraphics[width=\linewidth]{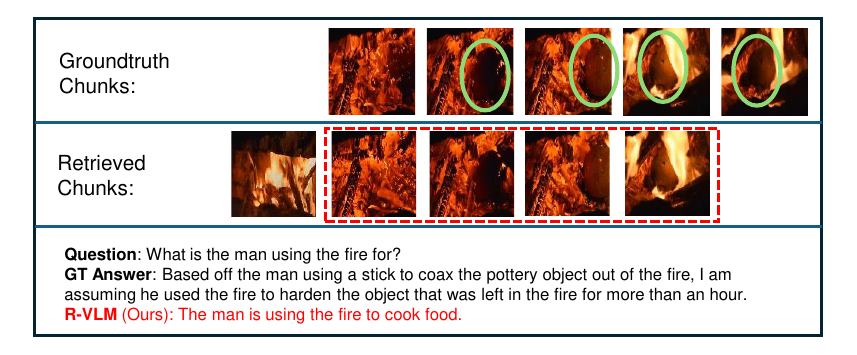} 
        \vspace{-1.5em}
        \caption{A failure case from WildQA. This is a video of a person firing art. Although our R-VLM retrieved the correct chunks, it gave the wrong answer of ``cook food”. This maybe caused by the visual ambiguity of the target object and the biases of the LLM.}
        \label{fig:bad_wildqa_1}  
    \end{subfigure}   
    \quad
    \begin{subfigure}{\mywidth\textwidth}  
        \centering
        \includegraphics[width=\linewidth]{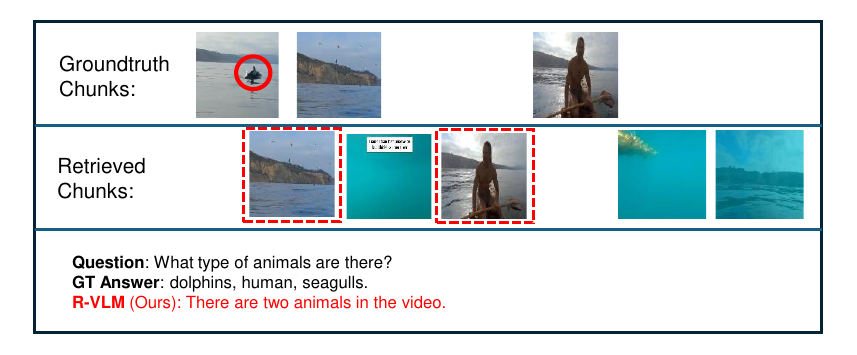} 
        \vspace{-1.5em}
        \caption{A failure case from WildQA.  ground-truth chunks correspond to the chunks where three types of animals present, namely dolphins (the first chunk), seagulls, and human. Our method only retrieved the seagull and human chunks, but missed the dolphin chunk. R-VLM provided wrong answer due to the imperfect retrieval and the unsatisfactory reasoning capability of the used LLM.}       
        \label{fig:bad_wildqa_2}  
    \end{subfigure}  
    \quad
    \begin{subfigure}{\mywidth\textwidth}  
        \centering
        \includegraphics[width=\linewidth]{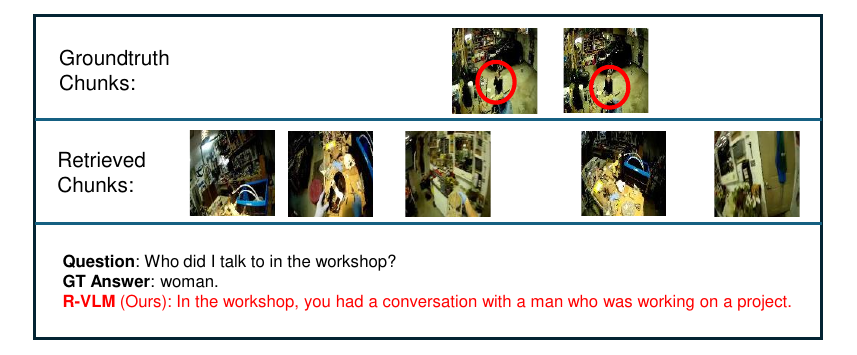} 
        \vspace{-1.5em}
        \caption{A failure case from QAEgo4D. Our method did not find the correct chunks. Therefore, large language model did not correctly answer the question and provided hallucinated answer.}      
        \label{fig:bad_qaego4d_1}  
    \end{subfigure}  
    \caption{Visualization of failure cases from WildQA and QAEgo4D.}  
    \label{fig:vis_failure_cases}  
\end{figure*} 

\vspace{-1mm}

\begin{figure*}[!htbp]
    \centering
    \begin{minipage}{\linewidth} 
        \centering
        \includegraphics[width=\linewidth]{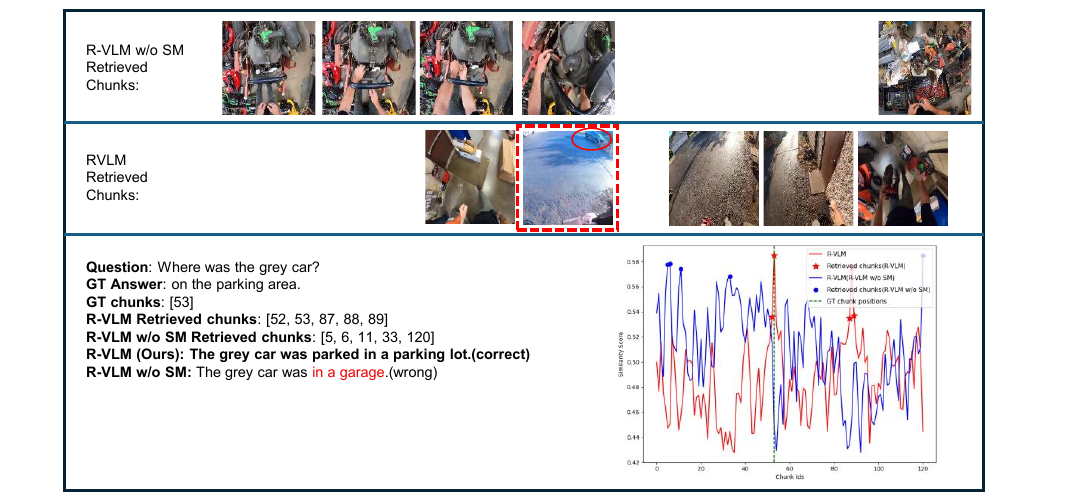}
        \vspace{-4mm}
        \caption{Visualization of video QA examples from QAEgo4D. For the curves, the red pointed stars indicate the video clip selected by R-VLM (red line), and the blue dots indicate the video clip selected by R-VLM w/o SM (blue line). The green dotted line indicates the ground truth question-related clips. After adding SM Loss training, our model R-VLM can capture the corresponding video clips more accurately.}
        \label{fig:good_qaego4d_1_smloss}
    \end{minipage}

    \begin{minipage}{\linewidth} 
        \centering
        \includegraphics[width=\linewidth]{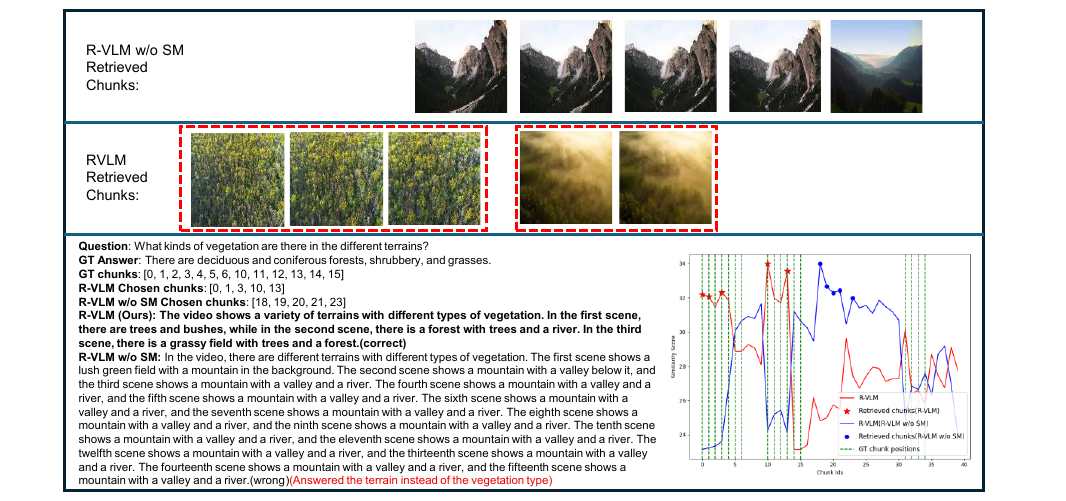}
        \vspace{-4mm}
        \caption{Visualization of video QA examples from WildQA. For the curves, the red pointed stars indicate the video clip selected by R-VLM (red line), and the blue dots indicate the video clip selected by R-VLM w/o SM (blue line). The green dotted line indicates the ground truth question-related clips. After adding SM Loss training, our model R-VLM can capture the corresponding video clips more accurately.}
        \label{fig:good_wildqa_3_smloss}
    \end{minipage}
\end{figure*}

\vspace{-15mm}
\subsection{Visualization Analysis}
\label{sec:morevis}

\noindent\textbf{Overall Visualization} We visualize examples from QAEgo4D in Fig.~\ref{fig:good_qaego4d_2}, Fig.~\ref{fig:good_qaego4d_3}, Fig.~\ref{fig:good_qaego4d_1}, and examples from WildQA in  Fig.~\ref{fig:good_wildqa_3}, Fig.~\ref{fig:good_wildqa_1}, Fig.~\ref{fig:good_wildqa_2}, with the following information.

1) The first row shows the video chunk examples by uniformly selecting 5 video chunks. 2) The second row shows the retrieved 5 chunks
(arranged in chronological order)
in our \ours{}. We mark the groudtruth chunks by red box. 3)
We show the curve of the learned similarity score (see Eq. (\ref{eq: similarity})) based on which the top $K$ chunks are selected.
The horizontal axis represents the indices of chunks and the vertical axis denotes the similarity score.
The groundtruth chunks and our retrieved chunks are also marked. 4) The question and answers from different models.

We can see that the predicted answers of our \ours{} is more accurate than \emph{Video-ChatGPT}, and \emph{Video-LLaMA}.  
For QAEgo4D, the average duration is about 8 minutes (120 video chunks), whereas
the groundtruth segments span only 2\% of the entire video on average.
It is difficult to hit groundtruth video segments by uniformly sampling $K(=5)$ chunks among 120 video chunks. Fig.~\ref{fig:good_qaego4d_2} shows that the fragments selected by uniform sampling are in general irrelevant to the problem. Our learnable retrieval can accurately find the segments where the answer lies in. Feeding the correct chunks to the LLM makes it possible to obtain the correct answer to the question. 

\vspace{2mm}

\noindent\textbf{Failure Case Visualization} In addition to successful cases, we present some failed cases in Fig.\ref{fig:vis_failure_cases}. There are two main cases of failure. One is that the retrieval did not select the correct video chunks. The other is that the retrieval correctly identified the correct video chunks, but the answer was wrong. For the latter cases, we think that more powerful vision feature extractor and LLMs would alleviate the problem.

\vspace{2mm}
\noindent\textbf{Visualization on Influence of SM Loss} We added visualizations showing similarity curves of chunk-question embeddings before and after SM loss training in Fig. \ref{fig:good_qaego4d_1_smloss} and Fig. \ref{fig:good_wildqa_3_smloss}. Note that for the curves, the red pointed stars indicate the video clip selected by R-VLM (red line), and the blue dots indicate the video clip selected by R-VLM w/o SM (blue line). The green dotted lines indicate the ground truth question-related clips. We can see that, in general, after adding SM Loss training, the model can capture the corresponding video clips more accurately.

\subsection{Human Evaluation Results}

We have conducted human studies in 100 samples of four test datasets on two models, the baseline scheme Video-ChatGPT, and our scheme R-VLM. We asked three people to blindly rate correct/wrong and scores and show the average evaluation results in Table \ref{tab:addlabel}. We can see that our method R-VLM achieves higher accuracy and score than Video-ChatGPT under human evaluation. The main trends from ChatGPT evaluation are consistent with our human evaluation, which indicates the reliableness of ChatGPT evaluation.

\begin{table}[htbp]
  \centering
  \caption{The results of human evaluation and ChatGPT evaluation on 100 samples, which are randomly selected from four test datasets.}
    \begin{tabular}{ccc}
    \toprule
    Evaluator & Video-ChatGPT & R-VLM (Ours) \\
    \midrule
    ChatGPT & 48.00/2.98 & 50.00/3.03 \\
    Human & 42.33/2.26 & 46.33/2.41 \\
    \bottomrule
    \end{tabular}%
  \label{tab:addlabel}%
\end{table}%

\section{Discussion}

\subsection{Rationale for Model Selection}
Here, we discuss the rationale for choosing specific models like MLP, CLIP image encoder, text encoder, and LLM.
\vspace{2mm}

\noindent\textbf{CLIP vision and text encoders} We use CLIP because it supplies robust, semantically aligned embeddings across modalities, thanks to its contrastive pretraining on 400M image-text pairs. This joint embedding space enables effective cross-modal alignment and efficient chunk scoring without expensive fine-tuning. CLIP’s encoders have been widely adopted in Video Language Modeling for VQA. 
\vspace{2mm}

\noindent\textbf{Large Language Model (LLM)} For understanding retrieved chunks and generating answers, LLMs offer strong reasoning and language capabilities. We chose LLaVA as our base model, which uses the popular LLM of LLaMA, enabling the process of natural language and visual context effectively.
\vspace{2mm}

\noindent\textbf{MLP for retrieval} A learnable two-layer MLP following the text encoder provides a lightweight and efficient way to score chunk relevance for retrieving question relevant video chunks.

\subsection{About Hour-long Videos}
We tested some videos from the popular long video benchmark LVBench (an average length of 68 minutes). We found that
(1) For question-video pairs with long certificate length (shortest video span that contains sufficient evidence to answer the question correctly), our model can capture the corresponding video chunks and generate answer correctly. For example, given the question: ``What does the girl use to find the person she is looking for in the video? (A) Mask (B) Tail (C) Phone (D) Car". Our model predicted the correct answer ``A". 
(2) For the pairs whose answers only exist in very short clips (1-2 chunks), since the average number of chunks is more than 1000, it becomes hard for our model to select the correct clips, making it difficult to give a correct answer. More efforts are still needed to deal with such hour-long videos and we leave as future work. 


\section{Limitation and Future Work}
\label{sec:limitations}
We have explored the retrieval-based solution for the long video understanding. We experimentally found that using $K=5$ selected chunks can generate superior performance. Sample adaptive $K$ should deliver optimal results but we leave this as future work. 

When the retrieval does not select the correct video chunks, our model will fail to generate correct answers as some failed cases in Fig.\ref{fig:vis_failure_cases}~(b) and (c). As a remedy, we could incorporate some global video tokens (e.g., like Video-ChatGPT) to assure the preservation of global information to alleviate this. We leave this as future work.

\end{document}